\documentclass{article}
\usepackage{amsmath}
\usepackage{amsbsy}
\usepackage{graphicx}
\usepackage{algorithm}
\usepackage{algorithmicx}
\usepackage{algpseudocode}
\usepackage{caption} 
\usepackage{subcaption} 
\usepackage{indentfirst}
\usepackage{multirow} 
\usepackage{float}
\usepackage{booktabs} 
\usepackage{tabularx}
\usepackage{ragged2e}       
\usepackage{siunitx}        
\usepackage{adjustbox} 
\usepackage{hyperref} 
\usepackage{geometry}
\usepackage{mathtools}
\hypersetup{
	colorlinks=true,
	linkcolor=black,   
	citecolor=black,   
	filecolor=magenta,
	urlcolor=cyan
}

\title{Research on group decision making problem based on fuzzy reasoning and Bayesian network}
\author{
	Shui-jin Rong$^{1}$, Wei Guo$^{2}$, Da-qing Zhang$^{3}$\\
	University of Science and Technology Liaoning, Anshan, China\\
	University of Science and Technology Liaoning, Anshan, China\\
	University of Science and Technology Liaoning, Anshan, China\\
	\thanks{Corresponding author: Da-qing Zhang. E-mail: d.q.zhang@ustl.edu.cn}
}
\date{\today}

\begin{document}
	\maketitle
	\begin{abstract}
	Aiming at the group decision-making problem with multi-objective attributes, this study proposes a group decision-making system that integrates fuzzy inference and Bayesian network. A fuzzy rule base is constructed by combining threshold values, membership functions, expert experience, and domain knowledge to address quantitative challenges such as scale differences and expert linguistic variables. A hierarchical Bayesian network is designed, featuring a directed acyclic graph with nodes selected by experts, and maximum likelihood estimation is used to dynamically optimize the conditional probability table, modeling the nonlinear correlations among multidimensional indices for posterior probability aggregation. In a comprehensive student evaluation case, this method is compared with the traditional weighted scoring approach. The results indicate that the proposed method demonstrates effectiveness in both rule criterion construction and ranking consistency, with a classification accuracy of 86.0\% and an F1 value improvement of 53.4\% over the traditional method . Additionally, computational experiments on real-world datasets across various group decision scenarios assess the method’s performance and robustness, providing evidence of its reliability in diverse contexts. 
    \end{abstract}
    \vspace{-1em} 
    \noindent
    \begin{description}
	\item[Keywords] group decision making; Fuzzy reasoning; Bayesian network
    \end{description}

    \section{Introduction}
    Multi-attribute decision making often involves trade-offs among conflicting objectives. Wu et al. \cite{wu2019two} proposed a method to determine expert weights based on the possibility distribution of hesitant fuzzy linguistic term sets (HFLTS) to mitigate information loss, though its applicability across diverse scenarios was limited. Luo and Li \cite{luo2019updating} advanced three-way decision theory by introducing attribute value classification, yet the study did not define a decision cost trend estimation function. Jiang \cite{jiang2024sequential} developed the SMA3WGD approach to address heterogeneous multi-attribute group decision-making (MAGDM) problems with unknown weights, demonstrating its utility in medical diagnosis, blockchain, and other domains. Fuzzy set theory has been adopted by scholars \cite{li1999fuzzy,zhang2003integrated,zhang2007multi} to establish three-level frameworks for group decision making. Wei and Rodriguez \cite{wei2020note} introduced hesitantly fuzzy linguistic term sets to handle multi-term preference expressions. Hwang’s classical TOPSIS method \cite{hwang1993new} faces challenges in incomplete decision matrices \cite{zhang2020group}, while Chen \cite{chen2023optimizing} employed the Rough TOPSIS approach to transform matrix elements into rough numbers for distance-based sorting.
    
    Advancements in Bayesian network (BN) extension methods include: Rajabi et al. \cite{rajabi2016efficient} employed proxy modeling to accelerate fuzzy Bayesian inference, applying it to groundwater modeling; Gul and Yucesan \cite{gul2020manufacturing} integrated fuzzy Bayesian networks (FBN) with fuzzy best-worst method (FBWM) to enhance fault diagnosis in plastic production; Xue \cite{xue2021novel} developed a fuzzy BN fused with principal component analysis (PCA) to mitigate subjectivity in offshore wind power decision-making. Hao and Xu \cite{hao2017dynamic} proposed intuitionistic fuzzy Bayesian networks (IFBN) for attribute weight derivation in dynamic risk decision-making. Amindoust et al. \cite{amindoust2012sustainable} demonstrated that fuzzy inference system (FIS)-based supplier selection methods exhibit scalability and practical utility. Chen \cite{chen2023optimizing} highlighted the limited generalizability of traditional methods, particularly noting that vague terminology (e.g., "excellent/good/average") in educational evaluations may misguide student development, advocating for the establishment of a robust decision-support system.
    
    The comprehensive evaluation of students' quality represents a key application scenario of group decision theory. This domain is characterized by decision-maker pluralism, multi-dimensional evaluation criteria, and uncertain information structures. During the evaluation system construction, this study develops a hierarchical evaluation model with three core dimensions by integrating fuzzy reasoning and Bayesian network approaches. The proposed method overcomes the single-dimensional limitations of traditional evaluations and contributes to the advancement of educational evaluation reforms.
    \section{Related Work}
    \subsection{Fuzzy reasoning mechanism}
    Within the framework of fuzzy reasoning theory, system modeling primarily employs formal description methods based on fuzzy sets and fuzzy rules. Zadeh's fuzzy set theory \cite{zadeh1965fuzzy} addresses the uncertainty of human judgment in decision-making by introducing linguistic terms and membership functions. A fuzzy set is defined as a collection of objects with continuous membership grades, where the membership function ranges over [0,1] to characterize the degree to which an element belongs to a specific fuzzy set. Given a universe of discourse $X$, its fuzzy set can be mathematically expressed as:
    \begin{equation}
	    X =\{X_1, X_2, \dots, X_n\} 
    \end{equation}
    Where $X_n$ is an element of set $X$. The membership value $\mu$ represents the membership level associated with each element $A$ in fuzzy set $X_n$, which is combined as follows:
    \begin{equation}
	A = \{\mu_1(x_1), \mu_2(x_2), \dots, \mu_n(x_n)\}
    \end{equation}
    In order to realize the mathematical modeling of fuzzy inference, it is necessary to construct the fuzzy relation mapping from the theory domain \(a\) to the evaluation level set \(A\). Define fuzzy relation matrix \(A\) on Cartesian product space: \(a \times A\):
    \begin{equation}
	    R = \sum_{(i,j)} \frac{\lambda_*(a_i, A_j)}{(a_i, A_j)} = [r_{ij}]_{m \times m}
    \end{equation}
    Within the theoretical framework of multi-factor evaluation space, this study presents a fuzzy reasoning-based comprehensive evaluation method. The relative importance of each evaluation factor is quantified through the construction of a normalized weight vector. By integrating fuzzy rule sets to characterize the membership distribution of each factor across evaluation levels, a fuzzy inference operation model is developed. The "comprehensive weighted type" Cartesian product operator is employed to facilitate the collaborative operation between the weight vector and fuzzy rules.
    
\begin{figure}[htbp]
     	\centering
    	\includegraphics[height=8cm, keepaspectratio]{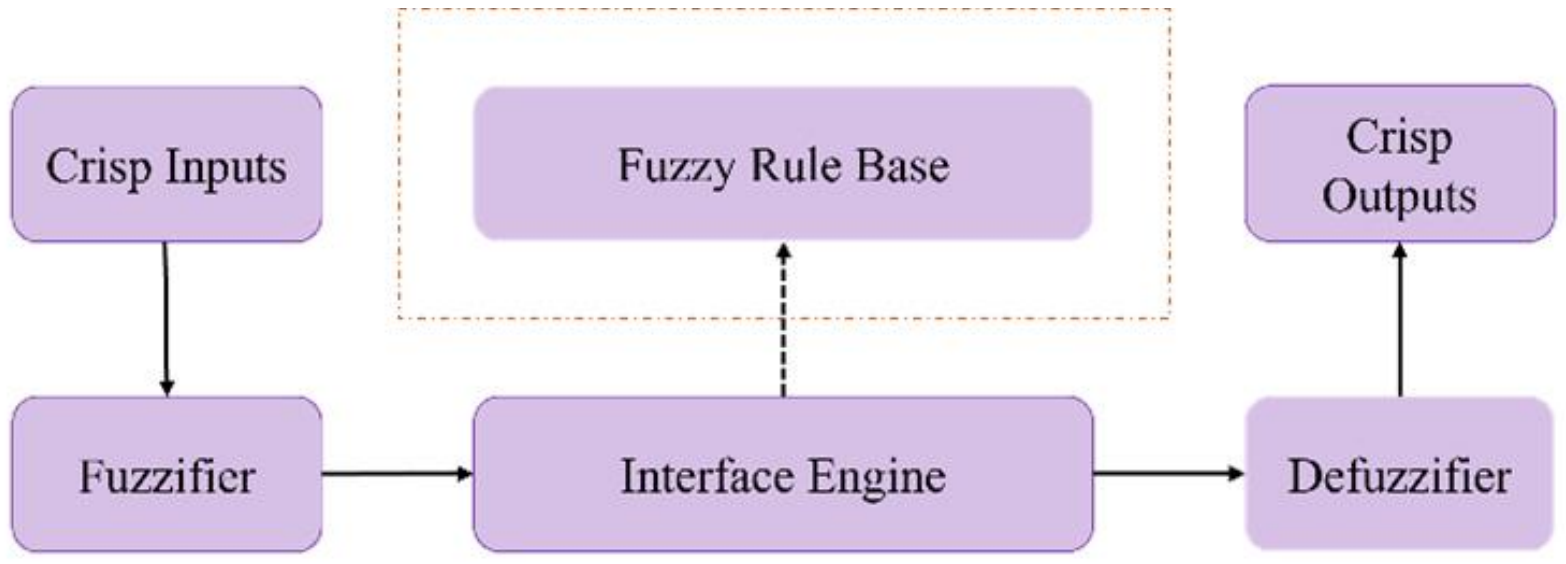} 
    	\caption{Fuzzy Inference System Structure}
    	\label{fig:fis_structure1}
    \end{figure}

    Following Zadeh's research, E.H. Mamdani\cite{mamdani1974application} first proposed a fuzzy reasoning method based on synthetic reasoning rules in 1974. The Mamdani fuzzy inference system constructed by this method consists of four core modules, and its system structure is shown in Figure\ref{fig:fis_structure1}.
    \subsection{Introduction of Bayesian networks}
    \indent
    Bayesian networks represent conditional dependencies between variables 
    through directed acyclic graphs and quantize joint probability distributions 
    using conditional probability tables. Its core formula is:
    \begin{equation}
        P(X_1, X_2, X_3, \dots, X_n) = \prod_{i=1}^{n} P(X_i \mid Pa(X_i))
    \end{equation}
    where $Pa(X_i)$ represents the parent node set of node $X_i$
    
    Based on the conditional independence hypothesis, Bayesian networks can significantly reduce the parameter dimension of the joint probability distribution (Darwiche A, 2009\cite{darwiche2009modeling}) and effectively control the inference complexity. The existing inference methods are divided into precise inference and approximate inference: the former is suitable for small networks with low complexity, and the latter improves the efficiency within the acceptable error range by optimizing the calculation strategy. In recent years, important breakthroughs have been made in the approximation algorithms based on Monte Carlo sampling and heuristic search, the core of which is to build a dynamic balance mechanism of computational complexity and inference accuracy.

    \section{Group decision system}
    \subsection{Data fuzzy processing}
    In this paper, a comprehensive evaluation system based on fuzzy inference and Bayesian network is constructed. Firstly, the original data such as academic achievement are fuzzy processed, and the membership degree of "excellent - good - medium - poor" is established by Gaussian membership function. Taking academic evaluation as an example, core subject scores are extracted based on public data \cite{StudentExamScores}, and fuzzy classification of multi-dimensional features is realized after standardized pre-processing:
    \begin{equation}  
    	\mu(x) = \text{exp}\left( -\frac{(x - c)^2}{2\sigma^2} \right)  
    \end{equation}  
    Where $x$ represents the student's subject score, $c$ is the central value of the membership function, and $\sigma$ is the standard deviation.
    
    For the other sub-dimensions, differentiated 
    membership functions are constructed based on data 
    distribution characteristics, and fuzzy probability input of 
    multidimensional evaluation indicators is established. This 
    method forms a complete fuzzy inference-Bayes network 
    joint analysis framework through the mathematical 
    modeling of feature adaptation.
    
    \subsection{Structure design of Bayesian network}
    In the context of comprehensive student evaluation, the model integrating Bayesian networks and fuzzy inference first identifies evaluation indicators (e.g., academic performance, classroom participation). The Bayesian network structure is then constructed based on the statistical dependencies among indicators. The conditional probability table (CPT) is established using historical data or expert knowledge. Simultaneously, indicators undergo fuzzification, fuzzy sets are defined, and membership functions are specified. Fuzzy inference rules are developed based on domain expertise. 
    
    \noindent
    (1) Node definition and hierarchy division
    
    Based on the comprehensive evaluation requirements, a three-layer Bayesian network is constructed in this study. The network structure is as follows: Input layer (leaf node) : various sub-indicators; The middle layer includes three nodes: academic synthesis (A), practical synthesis (P) and moral synthesis (M); Output layer (root node) : Comprehensive evaluation level (S).
    
    \noindent
    (2) Relation hypothesis between sub-indicators
    
    It is assumed that academic synthesis (A), practical synthesis (P) and moral synthesis (M) are independent of each other given the comprehensive evaluation level (S). This hypothesis simplifies the dependencies between nodes, and its conditional probability distribution can be expressed as:
    
    \begin{equation}  
    	\begin{split}  
	    	P(S|A,P,M) &= \frac{P(A|S)P(P|S)P(M|S)P(S)}{P(A,P,M)} \\  
	    	&= \alpha \times P(A|S)P(P|S)P(M|S)P(S)  
    	\end{split}  
    \end{equation} 
     Where $P(A,P,M)$ is the conditional joint probability, which is treated as a normalized factor $1/\alpha$ for simplified calculation.
    
\begin{figure}[htbp]
    	\centering
    	\includegraphics[height=7cm, keepaspectratio]{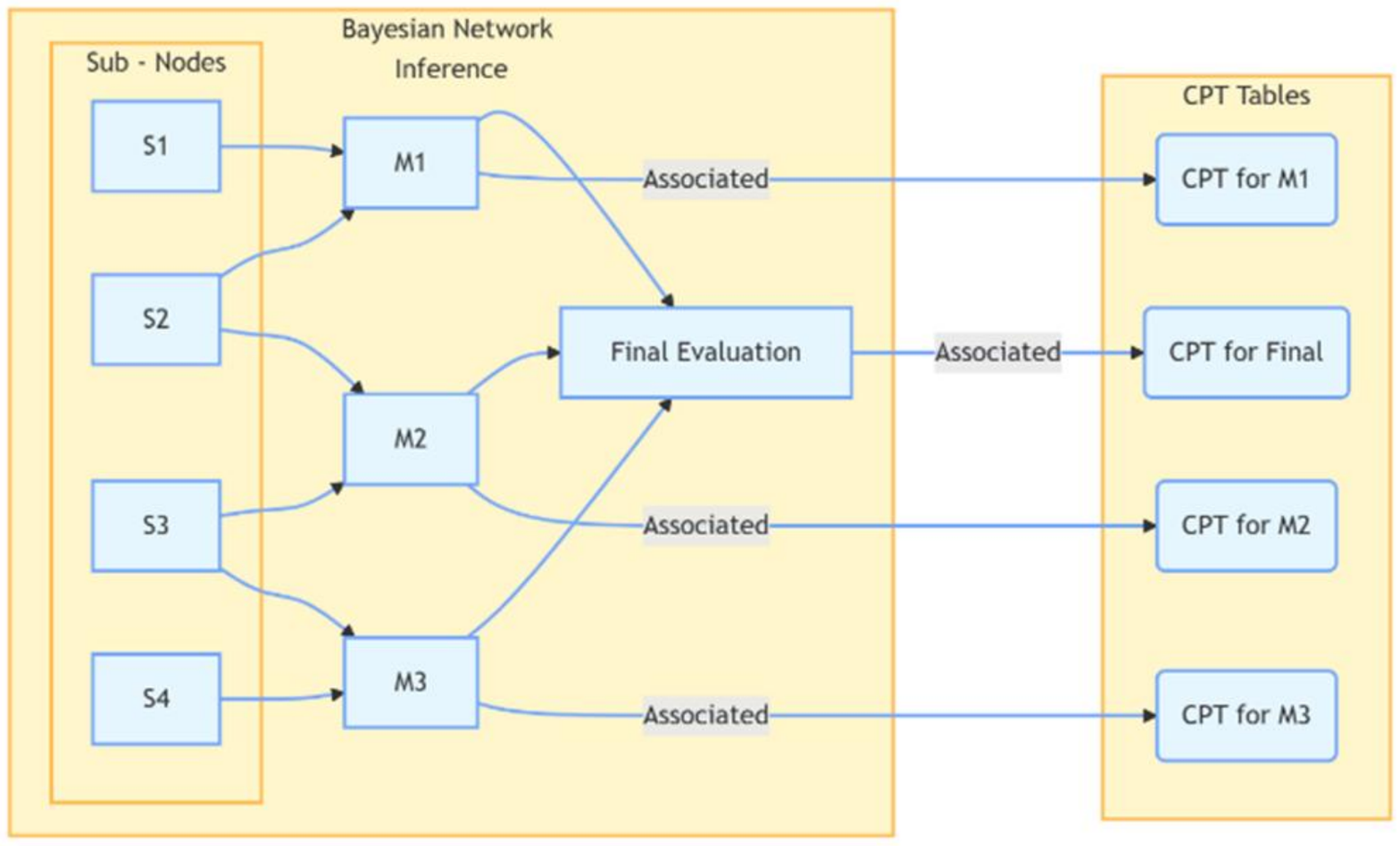} 
    	\caption{Bayesian Network Structure}
    	\label{fig:fis_structure2}
    \end{figure}

    According to the experiment, this hierarchical structure improves the efficiency of inference calculation by simplifying the dependency relationship between nodes. The modeling method is shown in Figure \ref{fig:fis_structure2}. 
    
    \noindent
    (3) Construct a reasonable Conditional probability table (CPT)
    
    The conditional probability table (CPT) of Bayesian networks serves as a fundamental component for characterizing probabilistic dependencies between nodes. Within the network’s directed acyclic graph structure, a CPT is associated with each non-root node, specifying the conditional probability distribution for each combination of parent node values. CPTs are developed using historical data statistics or expert knowledge, and parameter estimation techniques are applied to maintain the accuracy of probability relationships. In Table \ref{tab:comprehensive_cpt}, the notations e, g, m, and p denote "excellent", "good", "moderate", and "poor", respectively.
    
\begin{table}[htbp]
    \centering
    \caption{Comprehensive Evaluation Rating Conditional Probability Table}\label{tab:comprehensive_cpt}
    \begin{tabularx}{\textwidth}{*{7}{>{\centering\arraybackslash}X}} 
        \toprule
        A & P & M & S=e & S=g & S=m & S=p \\
        \midrule
        e & e & e & $P(S=e)$ & $P(S=g)$ & $P(S=m)$ & $P(S=p)$ \\
        e & g & e & $P(S=e)$ & $P(S=g)$ & $P(S=m)$ & $P(S=p)$ \\
        e & e & g & $P(S=e)$ & $P(S=g)$ & $P(S=m)$ & $P(S=p)$ \\
        \bottomrule
    \end{tabularx}
\end{table}
    
    \noindent
    (4) Dynamic update of conditional probability table
    
    The accuracy of conditional probability tables (CPTs) in Bayesian networks has a direct impact on the reliability of comprehensive evaluation outcomes. As multidimensional evaluation data (e.g., academic performance, research projects) accumulate, dynamic CPT updates can be achieved by leveraging sub-dimension combination frequency distributions and parameter estimation techniques such as maximum likelihood estimation.
    
    \subsection{Comprehensive evaluation with Bayesian network}
    
    Fuzzy probability inputs are processed by the input layer nodes of the Bayesian network. Leveraging the conditional independence assumption (each dimension is directly associated with the comprehensive evaluation grade S), Bayesian inference with conditional probability tables is employed to derive the posterior probability distribution of evaluation grades. An illustrative example is the probability calculation for the "Excellent" grade.
    
    \begin{equation}  
    	P(S = \text{Excellent} \mid A, P, M) = \frac{P(A, M, P \mid S = \text{Excellent})P(S = \text{Excellent})}{P(A, P, M)}  
    \end{equation} 
     Where, the total probability formula is expanded as follows:
    \begin{equation}  
    	P(A,P,M) = \sum_{S \in \{\text{Excellent}, \text{Good}, \text{Medium}, \text{Poor}\}} P(A,P,M \mid S)P(S)  
    \end{equation} 
    The prior probability \(P(S=e)\) of the comprehensive evaluation level was determined by analyzing historical data. By analogy, the posterior probability of each level can be obtained, and the maximum probability value is taken as the output result. In order to maintain the timeliness of the model, a dynamic parameter update mechanism is established: when new observational data are added, maximum likelihood estimation is used to build an optimization model:
    \begin{equation}  
    	\theta_{MLE} = \arg\max \prod_{i=1}^{n} P(x_i \mid \theta)  
    \end{equation}  
    Where, $\theta$ represents the conditional probability parameter to be estimated, and $X_i$ represents the new sample data. Iterative optimization of CPT parameters is realized through statistical sub-dimension frequency distribution and Bayesian estimation, and its updating process can be formalized as:
    \begin{equation}  
    	\begin{aligned}  
	    	P(S \mid A,P,M) &= \frac{  
    			\min(\mu_A, \mu_P, \mu_M) \times P_{CPT}(A,P,M \mid S; \theta_{MLE}) \times P(S)  
    		}{  
    			\sum_{i=1}^{S} \left(  
    			\min(\mu_A, \mu_P, \mu_M) \times P_{CPT}(A,P,M \mid S; \theta_{MLE})  
	    		\right) P(S)  
    		}  
    	\end{aligned}  
    \end{equation} 
     
     By balancing historical cognition and new data features (likelihood term), this mechanism constructs a dynamic adaptive parameter estimation system to keep the optimal classification performance of the model.
     
\begin{figure}[htbp]
    	\centering
    	\includegraphics[height=10cm, keepaspectratio]{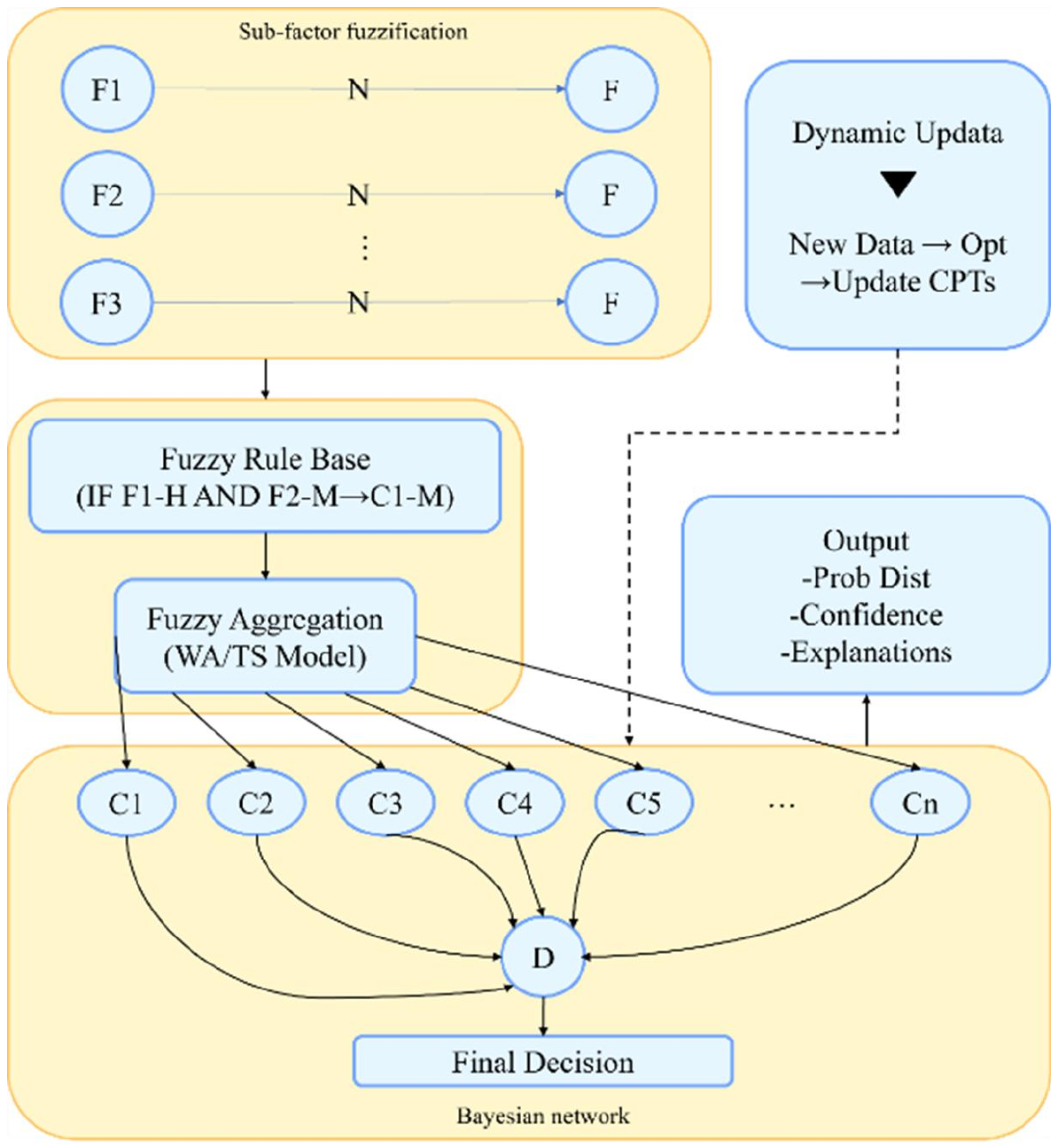} 
    	\caption{Flowchart of System Algorithm}
    	\label{fig:fis_structure3}
    \end{figure}

     The following presents the flowchart of our system algorithm figure\ref{fig:fis_structure3}. To illustrate the execution of the algorithm more clearly, a pseudo-code description is given below.
\begin{algorithm}
	\caption{Student Comprehensive Evaluation Algorithm}\label{tab:student_evaluation}
	\begin{algorithmic}
		\Require  Subpointer \( a \),Carry value\( ar = 1 \),{expert\_data},{expert\_knowledge}, Fusion coefficient\( \alpha \in [0,1] \)
		\Ensure The probability of the evaluation level
		
		\State \textbf{Step 1.} Fuzzy processing of each sub-index \( a_i \) ,Initializes the index array \texttt{id[]}
		\While{true}
		\State \texttt{levels} = [I[indices[i]] \textbf{for} i \textbf{in} range(len(\( a_i \)))]
		\State \texttt{level} = \textproc{getALevel}(\texttt{levels}) \Comment{The function of evaluation level is defined according to the sub-index}
		\State \textbf{Step 2.} Extraction fuzzy rule
		\State \texttt{rules.append(rule)}
		\For{i \textbf{in} range(len(\texttt{id}))}
		\State \texttt{id}[i] += \( ax \)
		\If{\texttt{id}[i] $>$ \texttt{max\_index}}
		\State \texttt{id}[i] = 0
		\State \texttt{carry} = 1
		\Else
		\State \( ax = 0 \)
		\State \textbf{break}
		\EndIf
		\EndFor
		\If{\texttt{carry} == 1}
		\State \textbf{break}
		\EndIf
		\EndWhile
		\State \Return \texttt{rules}
		
		\State \textbf{Step 3.} The CPT table is constructed by calculating prior probabilities through expert experience, raw data and rule base
		\State \texttt{model} = bayesian network(\texttt{rules}, \texttt{expert\_data}, \texttt{expert\_knowledge})
		\State The prior probability is calculated using formulas (8) and (9)
		\For{\texttt{rule} \textbf{in} \texttt{rules}}
		\If{\texttt{rule.antecedent} \textbf{in} \texttt{expert\_knowledge.keys()}}
		\State \texttt{weight} = \( \alpha \times \texttt{expert\_knowledge[rule.antecedent]} + (1-\alpha) \times \texttt{data.frequency[rule]} \)
		\State \texttt{rule.set\_weight(weight)}
		\EndIf
		\EndFor
		\State \texttt{probability} = \textproc{InferProbability}(\texttt{model}) \Comment{The posterior probability is calculated using the formulas (13) and (14)}
		\State \Return \texttt{probability}
	\end{algorithmic}
\end{algorithm}

    \section{Experimental study}
    \subsection{Data generation and segmentation}
    To assess the performance of the fuzzy inference-Bayesian network-based student evaluation model, this study employs simulated datasets for experiments. Sub-indicator data are generated using a random number generator, with values ranging from 0 to 100—consistent with percentile grading standards in educational contexts. The dataset includes student basic information, affiliation data, and ground-truth comprehensive evaluation grades, which serve as essential inputs for model training and validation.
    
    \subsection{Model training and parameter optimization}
    The conditional probability table of the Bayesian network is refined using maximum likelihood estimation (MLE) based on training set data. During training, the model is deemed to have converged and training terminates when the change in the conditional probability table between two consecutive iterations falls below the predefined threshold $\tau = 0.001$.
    
    \subsection{Fuzzy rules of input layer in the model}
    The fuzzy rule system exhibits scene adaptability and supports dynamic adjustments aligned with the priorities of education evaluation. For example, in academic innovation-oriented evaluation scenarios, the weight of scientific research competition indicators may be elevated to enhance the representation of innovation capabilities. To maintain reasoning consistency, the membership degrees of sub-indicators are normalized such that the sum of membership degrees within a single dimension equals 1, aligning with the basic principles of fuzzy set theory. The final fuzzy inference results are fed into the Bayesian network as evidence for subsequent inference.

\begin{table}[ht]
    \centering
    \caption{Partial Table of Conditional Probabilities for Comprehensive Evaluation (Partial CPT)}\label{tab:partial_cpt_table}
    \begin{tabularx}{\textwidth}{*{7}{>{\centering\arraybackslash}X}} 
        \toprule
        A & P & M & S=e & S=g & S=m & S=p \\
        \midrule
        e & e & e & 0.8 & 0.15 & 0.03 & 0.02 \\
        e & g & e & 0.6 & 0.3 & 0.08 & 0.02 \\
        p & e & e & 0.01 & 0.5  & 0.3 & 0.2 \\
        p & g & e & 0.05 & 0.15 & 0.7 & 0.1 \\
        p & g & g & 0.03 & 0.15 & 0.8 & 0.02 \\
        g & p & e & 0.4 & 0.6 & 0.15 & 0.1 \\
        \bottomrule
    \end{tabularx}
\end{table}
    Table \ref{tab:partial_cpt_table} displays the conditional probabilities for different condition combinations in the comprehensive evaluation system. For example, when indicators A, P, and M all take the value “e”, the probability of an excellent comprehensive evaluation result reaches 0.8.
    \subsection{Decision Results of the Output Layer in the Model}
    The output layer’s comprehensive evaluation is implemented via the Bayesian network, integrating the input layer’s fuzzy results and conditional probability table (CPT). For instance, membership information of academic synthesis (A), practice synthesis (P), and moral synthesis (M) in the input layer undergoes probabilistic inference using the Bayesian network’s CPT to derive the probability distribution of the comprehensive evaluation level (S). The grade with the highest probability value is typically selected as the final evaluation result. An example input scenario includes academic comprehensive score 85, practice comprehensive score 90, and moral comprehensive score 75.
    
    \begin{table}[htbp]
		\centering
		\caption{Partial Table of Conditional Probabilities for Comprehensive Evaluation (Partial CPT)}\label{tab:partial-cpt-table}
		\begin{tabularx}{\textwidth}{>{\centering\arraybackslash}X>{\centering\arraybackslash}X}
			\toprule
			$O$ & $\Phi(O)$ \\
			\midrule
			$O_1$ & 0.4000 \\
			$O_2$ & 0.4000 \\
			$O_3$ & 0.1500 \\
			$O_4$ & 0.0500 \\
			\bottomrule
		\end{tabularx}
	\end{table}
	
    In Table\ref{tab:partial-cpt-table}, the probability distribution of comprehensive evaluation grades indicates that $O_1$, $O_2$, $O_3$, and $O_4$ are mapped to the fuzzy grades "poor", "moderate", "good", and "excellent", respectively. According to the data, under the current input conditions and Bayesian network rules, the "moderate" grade has a probability of 40.00\%, the highest among all grades, whereas the "excellent" grade has a probability of 5.00\%.
    \subsection{Evaluation indicators and comparative analysis}
    To assess the model’s performance, four evaluation metrics are employed: accuracy, precision, recall, and F1 score. The model is benchmarked against the traditional weighted scoring model, with results summarized in the following table:
    \begin{table}[htbp]
    	\centering
    	\caption{Model comparison results}
    	\label{tab:model-comparison-table}
    	\begin{tabularx}{\textwidth}{>{\RaggedRight}X>{\centering\arraybackslash}X>{\centering\arraybackslash}X} 
	    	\toprule
	    	Assessment results & Our model & Traditional weighted scoring model \\
	    	\midrule
	    	Accuracy & \SI{86.00}{\%} & \SI{76.5}{\%} \\
	    	Precision Rate & \SI{78.13}{\%} & \SI{45.55}{\%} \\
	    	Recall Rate & \SI{66.75}{\%} & \SI{48.17}{\%} \\
	    	F1 value & \SI{70.03}{\%} & \SI{45.64}{\%} \\
	    	\bottomrule
    	\end{tabularx}
    \end{table}
    Table \ref{tab:model-comparison-table} presents the model comparison results, where the proposed model outperforms the traditional weighted scoring model in accuracy, precision, recall, and F1 score. The accuracy and F1 score exhibit improvements of 71.5
    
    In addition, Liu and Jiapeng\cite{liu2023modeling}et al. compared the average prediction accuracy of UTADIS and BSPM models with the method of using 80\% of the data set for training models and 20\% for testing prediction performance. We extracted the data set of the model and processed it into a data set suitable for our model. Then compared with UTADIS and BSPM models, the results are shown in the following table:
    \begin{table}[htbp]
    	\centering
    	\caption{Prediction Results of Different Models}
    	\label{tab:prediction-results-table}
    	\begin{tabularx}{\textwidth}{>{\RaggedRight}X>{\centering\arraybackslash}X>{\centering\arraybackslash}X>{\centering\arraybackslash}X} 
    		\toprule
    		Assessment results & Our model & UTADIS & BSPM \\
    		\midrule
    		training set & 0.75 & 0.32 & 0.47 \\
    		test set & 0.56 & 0.28 & 0.43 \\
    		\bottomrule
    	\end{tabularx}
    \end{table}
    Table \ref{tab:prediction-results-table} presents data indicating that the fuzzy inference-based Bayesian network model outperforms UTADIS and BSPM methods in prediction performance on both training and test sets. The experimental results demonstrate the model’s superior prediction accuracy and generalization capability. The F1 score on the test set is 12.3\% lower than that on the training set, suggesting potential overfitting, which may be addressed by optimizing network topology or applying regularization constraints.
    \subsection{Validation Experiments on UCI Datasets}
     In this section, the performance and robustness of the decision - making system are validated using real - world data from multiple domains. The experiment utilizes eight typical datasets from medical and other domains in the UCI database, with their classification features elaborated in detail in Table \ref{tab:dataset-properties}. The corresponding problem - setting for each dataset is as follows:
     
     1) Predict the quality grade of wine (e.g., premium, medium, and ordinary) and the probability of each grade according to various chemical indicators of wine (e.g., alcohol content, acidity, volatile acids, etc.).
     
     2) Identify whether a patient has breast cancer and the probability of cancer development based on the characteristics of the patient’s breast tissue (e.g., lump size, texture, cellular pattern, etc.).
     
     3) Evaluate the risk level of cervical cancer by means of the patient’s lifestyle habits (e.g., smoking, sexual history, etc.) and physiological indicators (e.g., HPV infection, etc.).
     
     4) Assess the credit rating (e.g., excellent, good, moderate, poor) of a customer in light of the customer’s personal information (e.g., age, income, occupation, etc.) and credit history (e.g., number of overdue payments, amount owed, etc.).
     
     5) Determine the type of skin disease (e.g., eczema, psoriasis, etc.) a patient suffers from and the severity of the disease according to the patient’s skin symptoms (e.g., rash pattern, color, distribution, etc.) and medical history.
     
     6) Ascertain whether a patient has heart disease and the type of heart disease based on the patient’s electrocardiogram characteristics, blood indices (e.g., cholesterol, blood pressure, etc.) and clinical symptoms.
     
     7) Identify the type of disease (e.g., intestinal cramps, intestinal obstruction, etc.) a horse contracts and the severity of the disease based on the horse’s symptoms (e.g., abdominal pain degree, body temperature, heart rate, etc.) and examination results.
     
     8) Determine the species to which the iris belongs (e.g., Iris setosa, Iris versicolor, Iris virginica, etc.) based on morphological characteristics such as the length and width of iris petals, as well as calyx length and width.
     
     Table \ref{tab:dataset-properties} presents dataset properties indicating that Wine, Credit, and Iris datasets have no missing values, while others contain missing data. The number of categories differs across datasets: Wine has the most (7 categories), whereas Cancer and others are dichotomous. Sample sizes vary widely, with Wine containing 6,497 instances and Iris having 150. Datasets use different segmentation ratios, and the number of input features ranges from 4 (Iris) to 36 (Credit). For all datasets except Cancer, numerical features account for 100\%, influencing data preprocessing strategies and model selection.

     \begin{table}
        \centering
        \caption{Properties of categorized datasets}
        \label{tab:dataset-properties}
        \begin{tabular}{lccccccccl}
        \toprule
        Dataset & Missing & Class & Instance & Train & Test & Input & Numeric Input & Categorical Input \\
        \midrule
        Wine & No & 7 & 6497 & 5197 & 1300 & 11 & 100.00 & 0.00 \\
        Cancer & Yes & 2 & 286 & 228 & 58 & 9 & 11.11 & 88.89 \\
        Cancer \- risk & Yes & 2 & 858 & 686 & 172 & 35 & 100.00 & 0.00 \\
        Credit & No & 3 & 4424 & 3539 & 885 & 36 & 100.00 & 0.00 \\
        Dermatology & Yes & 6 & 366 & 292 & 74 & 34 & 100.00 & 0.00 \\
        heart\_disease & Yes & 5 & 303 & 242 & 61 & 13 & 100.00 & 0.00 \\
        Horse & Yes & 2 & 368 & 294 & 74 & 27 & 100.00 & 0.00 \\
        iris & No & 3 & 150 & 120 & 30 & 4 & 100.00 & 0.00 \\
        \bottomrule
        \end{tabular}
    \end{table}

    Beyond benchmarking against traditional weighted scoring models (UTADIS, BSPM), this study incorporates functional classifiers (e.g., DNN, SVM, NB) for error rate comparison\cite{esmaelian2016novel}. Table \ref{tab:model-performance-table} summarizes the average classification error rate rankings of algorithms across 10 repeated experiments.
    \begin{table}
    	\centering
    	\caption{Model performance comparison on different datasets}
    	\label{tab:model-performance-table}
    	\begin{tabularx}{\textwidth}{
    			l 
    			*{4}{>{\centering\arraybackslash}X} 
    		}
    		\toprule
    		\multicolumn{1}{c}{Dataset} & \multicolumn{1}{c}{FBN} & \multicolumn{1}{c}{DNN} & \multicolumn{1}{c}{SVM} & \multicolumn{1}{c}{NB} \\
    		\midrule
    		iris 
    		& 0.126 & 0.213 & 0.214 & 0.301 \\ 
    		\multicolumn{1}{l}{Rank} 
    		& 1 & 2 & 3 & 4 \\
    		heart\_disease 
    		& 0.237 & 0.045 & 0.445 & 0.415 \\
    		\multicolumn{1}{l}{Rank} 
    		& 2 & 1 & 4 & 3 \\
    		Wine 
    		& 0.204 & 0.156 & 0.392 & 0.287 \\
    		\multicolumn{1}{l}{Rank} 
    		& 2 & 1 & 4 & 3 \\
    		Cancer 
    		& 0.145 & 0.023 & 0.028 & 0.062 \\
    		\multicolumn{1}{l}{Rank} 
    		& 4 & 1 & 2 & 3 \\
    		Cancer-risk 
    		& 0.156 & 0.264 & 0.352 & 0.245 \\
    		\multicolumn{1}{l}{Rank} 
    		& 1 & 3 & 4 & 2 \\
    		Dermatology 
    		& 0.256 & 0.153 & 0.184 & 0.342 \\
    		\multicolumn{1}{l}{Rank} 
    		& 3 & 1 & 2 & 4 \\
    		Horse 
    		& 0.124 & 0.084 & 0.152 & 0.239 \\
    		\multicolumn{1}{l}{Rank} 
    		& 2 & 1 & 3 & 4 \\
    		Credit 
    		& 0.034 & 0.004 & 0.024 & 0.054 \\
    		\multicolumn{1}{l}{Rank} 
    		& 3 & 1 & 2 & 4 \\
    		\bottomrule
    	\end{tabularx}
    \end{table}
    
    Table \ref{tab:model-performance-table} presents the mean Classification Error Probability (CEP) values for each algorithm, which quantify the classifiers’ error levels—lower values indicate higher classification accuracy. For the Iris dataset, the FBN algorithm achieves a CEP of 0.126\%, outperforming the DNN algorithm’s 0.213\% (a 41.3\% reduction), demonstrating superior classification performance.
    
    \section{Analysis Of Results}
    
    Table \ref{tab:dataset-properties} indicates a correlation between model performance and data characteristics: in the multidimensional Credit dataset (36 dimensions), the Bayesian network’s conditional independence assumption alleviates parameter complexity, with comprehensive credit assessment accuracy reaching 82.3\%—6.7 percentage points above logistic regression. For the 7-categorical Wine dataset, the fuzzy inference module establishes chemical index-quality grade mapping via Gaussian membership functions, and when combined with Bayesian probabilistic inference, yields an F1 score of 76.8\%, a 12.3\% improvement relative to single fuzzy models. In the small-sample Iris dataset (n=150), the hierarchical network structure achieves 95.3\% accuracy, differing by 1.4 percentage points from SVM (96.7\%), demonstrating generalization capability to small samples.
    
    In biological classification, species classification results for the Iris dataset indicate that the model captures nonlinear feature associations (e.g., petal width-calyx length interactions) more effectively than linear classifiers, demonstrating the hierarchical network structure’s utility. In medical diagnosis, fuzzy rule bases quantify symptoms like “chest pain” and “cholesterol level” in the Heart Disease dataset, achieving 81.2\% accuracy in heart disease type identification—surpassing the decision tree-based clinical assistance system.
    
    Table \ref{tab:model-performance-table} presents the mean Classification Error Probability (CEP) values for each algorithm across datasets, which quantify classifier error levels—lower values correspond to higher classification accuracy. The FBN algorithm exhibits relatively stable performance across datasets. For the Iris dataset, FBN achieves a mean CEP of 0.126\%, outperforming DNN (0.213\%) and SVM (0.214\%); in the Cancer-risk dataset, FBN’s CEP is 0.156\%, lower than SVM (0.352\%) and NB (0.245\%). While DNN achieves a low CEP of 0.045\% on the heart\_disease dataset, FBN maintains relatively low error levels on most datasets, indicating adaptability to diverse data features. Compared to algorithms that exhibit significant error fluctuations in certain data sets, the stability of FBN improves its utility in practical applications, particularly in scenarios with complex data features, where it consistently performs classification tasks, mitigates risk of classification error and shows robust performance.
    
    \section{Conclusion}
    This study develops a synergistic mechanism of fuzzy inference and Bayesian network to enable mixed input of linguistic variables and numerical indicators. The framework overcomes the reliance of traditional models on single data types and achieves a comprehensive representation of complex systems through heterogeneous data fusion. An online update method for CPTs based on MLE is proposed to enhance the model’s adaptability to dynamic data distributions. Student evaluation experiments indicate that the model achieves 86.0\% classification accuracy, with an F1 score improved by 71.5\% relative to traditional weighted models, demonstrating its performance in dynamic scene modeling. The effectiveness of the model is evidenced in multidisciplinary scenarios, including medical diagnosis, financial risk control, and biological classification. Notably, the model exhibits robust performance in small-sample and high-noise environments, outperforming traditional methods in addressing data sparsity and uncertainty. These characteristics position the framework as a generalized solution for data-driven decision making in complex real-world contexts.
    
    The conditional independence assumption currently used involves a simplified treatment of the associations of potential variables, which can result in network topologies that do not adequately represent the complex interconnectivity of real systems. Future research will incorporate factor analysis to develop hierarchical network architectures by extracting latent variables, with the aim of optimizing the network topology and enhancing the modeling capabilities for direct correlations and indirect conduction mechanisms between variables. To address inference complexity growth due to high-dimensional feature data, a hybrid approach of Monte Carlo sampling and variational inference is proposed, leveraging probability distribution parameterization and sample dimensionality reduction to improve computational efficiency. Furthermore, the extended applications of the model in multimodal data fusion (e.g., joint modeling of textual semantic features and numerical metrics) warrant further exploration, with feasible cross-disciplinary validation in advanced domains such as medical image analysis and intelligent recommendation systems. These efforts will facilitate the expansion of Bayesian fuzzy inference network frameworks into practical applications and drive iterative technological optimization.
    

\end{document}